%% file: main.tex
\documentclass{article}

\usepackage[utf8]{inputenc}
\usepackage{iclr2016_conference,times}
\usepackage{pgfplots}
\usepackage{amsmath}
\usepackage{amssymb}
\usepackage{sansmath}
\usepackage{bm}
\usepackage{booktabs}
\usepackage{pbox}
\usepackage{floatrow}
\usepackage{xcolor}
\usepackage{tcolorbox}
\usepackage{hyperref}
\usepackage{url}

\iclrfinalcopy

\title{A note on the evaluation of generative models}

\author{Lucas Theis\thanks{These authors contributed equally to this work.} \\
	University of Tübingen \\
	72072 Tübingen, Germany \\
	\texttt{lucas@bethgelab.org}
	\And
	Aäron van den Oord$^{*}$\thanks{Now at Google DeepMind.} \\
	Ghent University \\
	9000 Ghent, Belgium \\
	\texttt{aaron.vandenoord@ugent.be}
	\And
	Matthias Bethge \\
	University of Tübingen \\
	72072 Tübingen, Germany \\
	\texttt{matthias@bethgelab.org}
}

\newfloatcommand{capbtabbox}{table}[][0.48\textwidth]
\newfloatcommand{capbfigbox}{figure}[][0.48\textwidth]

\begin{document}
	\maketitle

	\begin{abstract}
	Probabilistic generative models can be used for
	compression, denoising, inpainting, texture synthesis, semi-supervised learning, unsupervised feature
	learning, and other tasks. Given this wide range of applications, it is not surprising that a lot of heterogeneity exists in the way
	these models are formulated, trained, and evaluated. As a consequence, direct comparison between
	models is often difficult. This article reviews mostly known
	but often underappreciated properties relating to the evaluation and interpretation of
	generative models with a focus on image models.
	In particular, we show that three of the currently most commonly used
	criteria---average log-likelihood, Parzen window estimates, and visual fidelity of
	samples---are largely independent of each other when the data is high-dimensional. Good performance with
	respect to one criterion therefore need not imply good performance with respect to the other
	criteria.
	Our results show that extrapolation from one criterion to another is not warranted and
	generative models need to be evaluated directly with respect to the application(s) they were
	intended for. In addition, we provide examples demonstrating that Parzen window estimates should
	generally be avoided.
	\end{abstract}

	\section{Introduction}
		Generative models have many applications and can be evaluated in many ways.
		For density estimation and related tasks, log-likelihood (or equivalently Kullback-Leibler divergence) has
		been the de-facto standard for training and evaluating generative models. However, the
		likelihood of many interesting models is computationally intractable.
		For example, the normalization constant of unnormalized energy-based models is generally
		difficult to compute, and latent-variable models often require us to solve complex integrals to compute the likelihood.
		These models may still be trained with respect to a different objective that is more or less related to log-likelihood,
		such as contrastive divergence \citep{Hinton:2002}, score matching
		\citep{Hyvaerinen:2005}, lower bounds on the log-likelihood \citep{Bishop:2006}, noise-contrastive estimation \citep{Gutmann:2010},
		probability flow \citep{Sohl-Dickstein:2011}, maximum mean discrepancy (MMD)
		\citep{Gretton:2007,Li:2015}, or approximations to the Jensen-Shannon divergence (JSD)
		\citep{Goodfellow:2014}.

		For computational reasons, generative models are also often compared in terms of
		properties more readily accessible than likelihood, even when the task is density estimation. Examples include visualizations of model samples,
		interpretations of model parameters \citep{Hyvarinen:2009}, Parzen window estimates of the
		model's log-likelihood \citep{Breuleux:2009}, and evaluations of model performance in surrogate tasks such as denoising or missing value imputation.

		In this paper, we look at some of the implications of choosing certain training and evaluation
		criteria. We first show that training objectives such as JSD and MMD can result in
		very different optima than log-likelihood.
		We then discuss the relationship between log-likelihood, classification
		performance, visual fidelity of samples and Parzen window estimates. We show that
		good or bad performance with respect to one metric is no guarantee of good or bad
		performance with respect to the other metrics. In particular, we show that
		the quality of samples is generally uninformative about the likelihood and vice versa, and that Parzen
		window estimates seem to favor models with neither good likelihood nor samples of highest
		possible quality. Using Parzen window estimates as a criterion, a simple model based on $k$-means outperforms the true
		distribution of the data.



	\section{Training of generative models}
		\begin{figure}[t]
			\centering
			\input{figures/metrics.tex}
			\caption{An isotropic Gaussian distribution was fit to data drawn from a mixture of
				Gaussians by either minimizing Kullback-Leibler divergence (KLD), maximum mean
				discrepancy (MMD), or Jensen-Shannon divergence (JSD). The different fits demonstrate
				different tradeoffs made by the three measures of distance between distributions.}
			\label{fig:metrics}
		\end{figure}
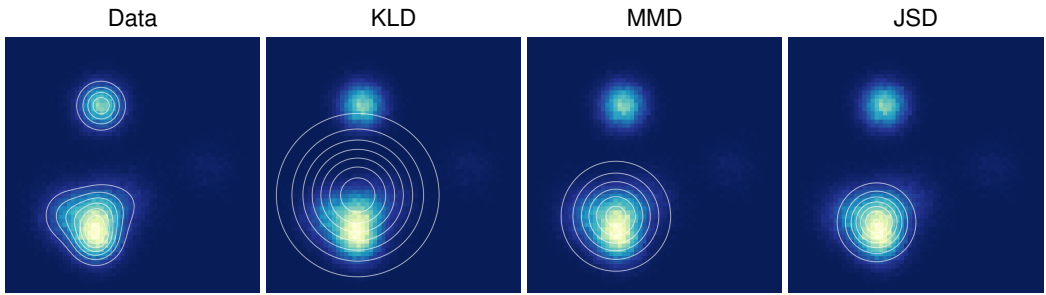

		Many objective functions and training procedures have been proposed for optimizing generative
		models. The motivation for introducing new training methods is
		typically the wish to fit probabilistic models with computationally intractable likelihoods,
		rendering direct maximum likelihood learning impractical. Most of
		the available training procedures are consistent in the sense that if the data
		is drawn from a model distribution, then this model distribution
		will be optimal under the training objective in the limit of an infinite number of training examples.
		That is, if the model is correct, and for extremely large amounts of data, all of these methods
		will produce the same result.
		However, when there is a mismatch between the data distribution and the model,
		different objective functions can lead to very different results.

		Figure~\ref{fig:metrics} illustrates this on a simple toy example where an
		isotropic Gaussian distribution has been fit to a mixture of Gaussians by minimizing various measures of distance.
		Maximum mean discrepancy (MMD) has been used with generative moment matching networks \citep{Li:2015,Dziugaite:2015}
		and Jensen-Shannon divergence (JSD) has connections to the objective function optimized by
		generative adversarial networks \citep{Goodfellow:2014} (see box for a definition).
		Minimizing MMD or JSD yields a Gaussian which fits one mode well, but which ignores other parts of the data.
		On the other hand, maximizing average log-likelihood or equivalently minimizing Kullback-Leibler divergence (KLD) avoids assigning extremely
		small probability to any data point but assigns a lot of probability mass to non-data regions.

		Understanding the trade-offs between different measures is important for several reasons. First,
		different applications require different trade-offs, and we want to choose the right metric
		for a given application. Assigning sufficient probability to all
		plausible images is important for compression, but it may be enough to generate a single
		plausible example in certain image reconstruction applications
		\cite[e.g.,][]{Hays:2007}.
		Second, a better understanding of the trade-offs allows us to better interpret and relate
		empirical findings.
		Generative image models are often assessed based on the visual fidelity of generated samples \cite[e.g.,][]{Goodfellow:2014,Gregor:2015,Denton:2015,Li:2015}.
		Figure~\ref{fig:metrics} suggests that a model optimized with respect to KLD is more likely
		to produce atypical samples than the same model optimized with respect to one of the
		other two measures. That is, plausible samples---in the sense of having large density
		under the target distribution---are not necessarily an indication of a good
		density model as measured by KLD, but may be expected when optimizing JSD.

		\begin{figure}[t]
			\begin{tcolorbox}[
					colback=gray!20,
					colframe=gray!20,
					arc=0pt,
					outer arc=0pt
				]
				\textbf{MMD} \citep{Gretton:2007} is defined as,
				\begin{align}
				\text{MMD}[p, q] = \left(
					\text{E}_{p,q}[k(\mathbf{x}, \mathbf{x}') - 2 k(\mathbf{x}, \mathbf{y}) + k(\mathbf{y}, \mathbf{y'})]
					\right)^\frac{1}{2},
				\end{align}
				where $\mathbf{x}, \mathbf{x}'$ are indepent and distributed according to the data
				distribution $p$, and $\mathbf{y}, \mathbf{y}'$ are independently distributed according to
				the model distribution $q$. We followed the approach of \citet{Li:2015}, optimizing an
				empirical estimate of MMD and using a mixture of Gaussian kernels with various bandwidths for $k$.
				\\
				\\
				\textbf{JSD} is defined as
				\begin{align}
					\text{JSD}[p, q] = \frac{1}{2} \text{KLD}[p \mid\mid m] + \frac{1}{2} \text{KLD}[q \mid\mid m],
				\end{align}
				where $m = (p + q) / 2$ is an equal mixture of distributions $p$ and $q$. We optimized JSD
				directly using the data density, which is generally not possible in practice where we only
				have access to samples from the data distribution. In this case, generative
				adversarial networks (GANs) may be used to approximately optimize JSD, although
				in practical applications the objective function optimized by GANs can be very
				different from JSD. Parameters were initialized at the maximum likelihood solution in all cases,
				but the same optimum was consistently found using random initializations.
			\end{tcolorbox}
		\end{figure}

	\section{Evaluation of generative models}
		Just as choosing the right training method is important for achieving good performance in a
		given application, so is choosing the right evaluation metric for drawing the right
		conclusions. In the following, we first continue to discuss the relationship between average
		log-likelihood and the visual appearance of model samples. 
		
		Model samples can be a useful diagnostic tool, often allowing us to build an intuition for why
		a model might fail and how it could be improved.
		However, qualitative as well as quantitative analyses based on model samples can be misleading
		about a model's density estimation performance, as well as the probabilistic model's
		performance in applications other than image synthesis.
		Below we summarize a few examples demonstrating this.

		\subsection{Log-likelihood}
		\label{sec:log-likelihood}
		Average log-likelihood is widely considered as the default measure for quantifying generative image
		modeling performance. However, care needs to be taken to ensure that the numbers measured
		are meaningful. While natural images are typically stored using 8-bit integers, they are
		often modeled using densities, i.e., an image is treated as an instance of a continuous random variable.
		Since the discrete data distribution has differential entropy of negative infinity, this can lead to arbitrary high likelihoods even
		on test data. To avoid this case, it is becoming best practice to add
		real-valued noise to the integer pixel values to dequantize the data
		\cite[e.g.,][]{Uria:2013,VanDenOord:2014,Theis:2015}.

		If we add the right amount of \emph{uniform} noise, the log-likelihood of the continuous
		model on the dequantized data is closely related to the log-likelihood of a discrete model on the discrete data.
		Maximizing the log-likelihood on the continuous data
		also optimizes the log-likelihood of the discrete model on the original data. This can be seen as follows.

		Consider images $\mathbf{x} \in \{ 0, ..., 255 \}^D$
		with a discrete probability distribution $P(\mathbf{x})$,
		uniform noise $\mathbf{u} \in \left[0, 1\right[^D$, and noisy data $\mathbf{y} = \mathbf{x} + \mathbf{u}$. If
		$p$ refers to the noisy data density and $q$ refers to the model density, then we have for the
		average log-likelihood:
		\newpage
		\begin{align}
			\int p(\mathbf{y}) \log q(\mathbf{y}) \, d\mathbf{y}
			\label{eq:loglik}
			&= \sum_{\mathbf{x}} P(\mathbf{x}) \int_{\left[0, 1\right[^D} \log q(\mathbf{x} + \mathbf{u}) \, d\mathbf{u} \\
			&\leq \sum_{\mathbf{x}} P(\mathbf{x}) \log \int_{\left[0, 1\right[^D} q(\mathbf{x} + \mathbf{u}) \, d\mathbf{u} \\
			&= \sum_{\mathbf{x}} P(\mathbf{x}) \log Q(\mathbf{x}),
			\label{eq:compression}
		\end{align}
		where the second step follows from Jensen's inequality and we have defined
		\begin{align}
			Q(\mathbf{x}) = \int_{\left[0, 1\right[^D} q(\mathbf{x} + \mathbf{u}) \, d\mathbf{u}
		\end{align}
		for $\mathbf{x} \in \mathbb{Z}^D$. The left-hand side in Equation~\ref{eq:loglik} is the expected log-likelihood
		which would be estimated in a typical benchmark. The right-hand side is the log-likelihood
		of the probability mass function $Q$ on the original discrete-valued image
		data.
		The negative of this log-likelihood is equivalent to the average number of bits (assuming base-2 logarithm)
		required to losslessly compress the discrete data with an
		entropy coding scheme optimized for~$Q$~\citep{Shannon:2001}.

		\subsubsection*{Semi-supervised learning}
		A second motivation for using log-likelihood comes from semi-supervised learning.
		Consider a dataset consisting of images $\mathcal{X}$ and corresponding labels $\mathcal{Y}$ for some but not necessarily all
		of the images. In classification, we are interested in the prediction of a class
		label $y$ for a previously unseen query image $\mathbf{x}$. For a given model relating
		$\mathbf{x}$, $y$, and parameters $\bm{\theta}$, the
		only correct way to infer the distribution over $y$---from a Bayesian point of view
		---is to integrate out the parameters \cite[e.g.,][]{Lasserre:2006},
		\begin{align}
			p(y \mid \mathbf{x}, \mathcal{X}, \mathcal{Y})
			= \int p(\bm{\theta} \mid \mathcal{X}, \mathcal{Y}) p(y \mid \mathbf{x}, \bm{\theta}) \, d\bm{\theta}.
		\end{align}
		With sufficient data and under certain assumptions, the above
		integral will be close to $p(y \mid \mathbf{x}, \bm{\hat\theta}_\text{MAP})$, where
		\begin{align}
			\bm{\hat\theta}_\text{MAP}
			&= \text{argmax}_{\bm{\theta}} \, p(\bm{\theta} \mid \mathcal{X}, \mathcal{Y}) \\
			\label{eq:map}
			&= \text{argmax}_{\bm{\theta}} \, \left[ \log p(\bm{\theta}) + \log p(\mathcal{X} \mid \bm{\theta}) + \log p(\mathcal{Y} \mid \mathcal{X}, \bm{\theta}) \right].
		\end{align}
		When no training labels are given, i.e., in the unsupervised setting, and for a uniform prior over
		parameters, it is therefore natural to try to optimize the log-likelihood, $\log p(\mathcal{X} \mid \bm{\theta})$.

		In practice, this approach might fail because of a mismatch between the model and the data,
		because of an inability to solve Equation~\ref{eq:map}, or because of overfitting induced by
		the MAP approximation. These issues can be addressed by better image models
		\citep[e.g.,][]{Kingma:2014}, better optimization and inference procedures, or a more Bayesian treatment of the parameters
		\citep[e.g.,][]{Lacoste-Julien:2011,Welling:2011}.

		\subsection{Samples and log-likelihood}
		For many interesting models, average log-likelihood is difficult to evaluate or even
		approximate. For some of these models at least, generating samples is a lot easier. It would
		therefore be useful if we could use generated samples to infer something about a model's log-likelihood.
		This approach is also intuitive given that a model with zero KL divergence will
		produce perfect samples, and visual inspection can work well in low dimensions for assessing
		a model's fit to data. Unfortunately these intuitions can be misleading
		when the image dimensionality is high. A model can have poor log-likelihood and produce
		great samples, or have great log-likelihood and produce poor samples.

		\subsubsection*{Poor log-likelihood and great samples}
		A simple lookup table storing enough training images will generate convincing looking images
		but will have poor average log-likelihood on unseen test data. Somewhat more generally we might
		consider a mixture of Gaussian distributions,
		\begin{align}
			q(\mathbf{x}) = \frac{1}{N} \sum_n \mathcal{N}(\mathbf{x}; \mathbf{x}_n, \varepsilon^2\mathbf{I}),
			\label{eq:lookup_table}
		\end{align}
		where the means $\mathbf{x}_n$ are either training images or a number of plausible images
		derived from the training set (e.g., using a set of image transformations). If $\varepsilon$ is small enough such that the Gaussian noise becomes imperceptible, this
		model will generate great samples but will still have very poor log-likelihood. This shows that
		plausible samples are clearly \textit{not sufficient} for a good log-likelihood.

		\citet{Gerhard:2013} empirically found a correlation between some models'
		log-likelihoods and their samples' ability to fool human observers into thinking they were
		extracted from real images. However, the image patches were small and all models used in
		the study were optimized to minimize KLD. The correlation between log-likelihood and sample
		quality may disappear, for example, when considering models optimized for different
		objective functions or already when considering a different set of models.

		\subsubsection*{Great log-likelihood and poor samples}
		Perhaps surprisingly, the ability to produce plausible samples is not only not
		sufficient, but also \textit{not necessary} for high likelihood as a simple argument by
		\citet{VanDenOord:2015} shows:
		Assume $p$ is the density of a model for $d$ dimensional
		data $\mathbf{x}$ which performs arbitrarily well with respect to average log-likelihood
		and $q$ corresponds to some bad model (e.g., white noise).
		Then samples generated by the mixture model
		\begin{align}
			0.01 p(\mathbf{x}) + 0.99 q(\mathbf{x})
			\label{eq:mix}
		\end{align}
		will come from the poor model 99\% of the time. Yet the log-likelihood per
		pixel will hardly change if $d$ is large:
		\begin{align}
			\log\left[ 0.01 p(\mathbf{x}) + 0.99 q(\mathbf{x}) \right]
			\geq \log\left[ 0.01 p(\mathbf{x}) \right]
			= \log p(\mathbf{x}) - \log 100
		\end{align}
		For high-dimensional data, $\log p(\mathbf{x})$ will be proportional to
		$d$ while $\log 100$ stays constant. For instance, already for the 32 by 32 images found
		in the CIFAR-10 dataset the difference between log-likelihoods of different models can be in
		the thousands, while $\log(100)$ is only about 4.61 nats \citep{VanDenOord:2015}. This shows
		that a model can have large average log-likelihood but generate very poor samples.

		\subsubsection*{Good log-likelihood and great samples}
		Note that we could have also chosen $q$ (Equation~\ref{eq:mix}) such that it reproduces training examples,
		e.g., by choosing $q$ as in Equation~\ref{eq:lookup_table}.
		In this case, the mixture model would generate samples indistinguishable from real images 99\% of the time
		while the log-likelihood would again only change by at most 4.61 nats.
		This shows that any model can be turned into a model which produces realistic samples at little expense to
		its log-likelihood. Log-likelihood and visual appearance of samples are therefore largely independent.

		\subsection{Samples and applications}
		\label{sec:applications}
		One might conclude that something must be wrong with log-likelihood if it does not care about a model's
		ability to generate plausible samples. However, note that the mixture model in Equation~\ref{eq:mix}
		might also still work very well in applications. While $q$ is much more likely a priori,
		$p$ is going to be much more likely a posteriori in tasks like inpainting, denoising, or
		classification. Consider prediction of a quantity $y$ representing, for example, a class label or missing
		pixels. A model with joint distribution
		\begin{align}
			0.01 p(\mathbf{x}) p(y \mid \mathbf{x}) + 0.99 q(\mathbf{x}) q(y \mid \mathbf{x})
		\end{align}
		may again generate poor samples 99\% of the time. For a given fixed $\mathbf{x}$, the posterior
		over $y$ will be a mixture
		\begin{align}
			\alpha p(y \mid \mathbf{x}) + (1 - \alpha) q(y \mid \mathbf{x}),
		\end{align}
		where a few simple calculations show that
		\begin{align}
			\alpha = \sigma\left( \ln p(\mathbf{x}) - \ln q(\mathbf{x}) - \ln 99 \right)
		\end{align}
		and $\sigma$ is the sigmoidal logistic function.
		Since we assume that $p$ is a good model, $q$ is a poor model, and $\mathbf{x}$
		is high-dimensional, we have
		\begin{align}
			\ln p(\mathbf{x}) \gg \ln q(\mathbf{x}) + \ln 99
		\end{align}
		and therefore $\alpha \approx 1$. That is, mixing with $q$ has hardly changed the posterior
		over $y$. While the samples are dominated by $q$, the classification performance is
		dominated by $p$. This shows that high visual fidelity of samples is generally not necessary for
		achieving good performance in applications.

		\subsection{Evaluation based on samples and nearest neighbors}
		A qualitative assessment based on samples can be biased towards models which overfit
		\citep{Breuleux:2009}. To detect overfitting to the training data, it is common to show samples next to nearest
		neighbors from the training set. In the following, we highlight two limitations of this
		approach and argue that it is unfit to detect any but the starkest forms of overfitting.

		\begin{figure}[t]
			\centering
			\input{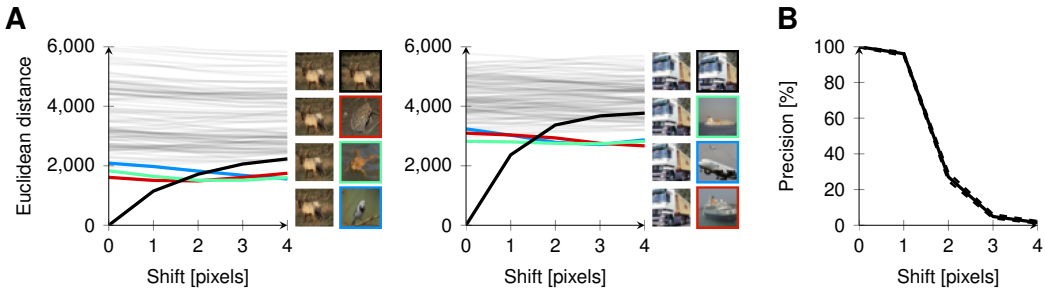}
			\caption{\textbf{A:} Two examples demonstrating that small changes of an image can
				lead to large changes in Euclidean distance affecting the choice of nearest neighbor.
				The images shown represent the query image shifted by between 1 and 4 pixels (left column,
				top to bottom), and the corresponding nearest neighbor from the training set (right
				column). The gray lines indicate Euclidean distance of the query image to 100 randomly
				picked images from the training set.
				\textbf{B:} Fraction of query images assigned to the correct training image. The
				average was estimated from 1,000 images. Dashed lines indicate a 90\% confidence
				interval.}
			\label{fig:samples}
		\end{figure}

		Nearest neighbors are typically determined based on Euclidean distance.
		But already perceptually small changes can lead to large changes in Euclidean distance,
		as is well known in the psychophysics literature \cite[e.g.,][]{Wang:2009}. To illustrate
		this property, we used the top-left 28 by 28 pixels of each image from the 50,000 training images of the
		CIFAR-10 dataset. We then shifted this 28 by 28 window one pixel down and one pixel to the
		right and extracted another set of images. We repeated this 4 times, giving us
		4 sets of images which are increasingly different from the training set.
		Figure~\ref{fig:samples}A shows nearest neighbors of corresponding images from the query
		set. Although the images have hardly changed visually, a shift by only two pixels already
		caused a different nearest neighbor. The plot also shows Euclidean distances to 100
		randomly picked images from the training set. Note that with a bigger dataset, a switch to
		a different nearest neighbor becomes more likely. Figure~\ref{fig:samples}B shows the
		fraction of query images assigned to the correct training image in our example.
		A model which stores transformed training images can trivially pass the nearest-neighbor overfitting test.
		This problem can be alleviated by choosing nearest neighbors based on perceptual metrics,
		and by showing more than one nearest neighbor.

		A second problem concerns the entropy of the model distribution and is
		harder to address. There are different ways a model can overfit. Even when overfitting,
		most models will not reproduce perfect or trivially transformed copies of the training data. In this case, no distance metric will find
		a close match in the training set. A model which overfits might still never generate a plausible image or
		might only be able to generate a small fraction of all plausible images (e.g.,
		a model as in Equation~\ref{eq:lookup_table} where instead of training images we
		store several transformed versions of the training images, or a model which only describes
		data in a lower-dimensional subspace). Because the number of images we
		can process is vanishingly small compared to the vast number of possible images,
		we would not be able to detect this by looking at samples from the model.

		\subsection{Evaluation based on Parzen window estimates}
		When log-likelihoods are unavailable, a common alternative is to use Parzen window
		estimates. Here, samples are generated from the model and used to construct a tractable
		model, typically a kernel density estimator with Gaussian kernel. A test log-likelihood is
		then evaluated under this model and used as a proxy for the true model's
		log-likelihood \citep{Breuleux:2009}. \citet{Breuleux:2009} suggested to fit the Parzen
		windows on both samples and training data, and to use at least as many samples as there are
		images in the training set. Following \cite{Bengio:2013a}, Parzen windows are in practice commonly fit to
		only 10,000 samples \cite[e.g.,][]{Bengio:2013b,Goodfellow:2014,Li:2015,Sohl-Dickstein:2015}.
		But even for a large number of samples Parzen window estimates generally do not come close
		to a model's true log-likelihood when the data dimensionality is high.
		In Figure~\ref{fig:parzen_windows} we plot Parzen window estimates for a multivariate Gaussian
		distribution fit to small CIFAR-10 image patches (of size 6 by 6).
		We added uniform noise to the data (as explained in Section~\ref{sec:log-likelihood}) and rescaled between 0 and 1.
		As we can see, a completely infeasible number of samples would be needed to
		get close to the actual log-likelihood even for this small scale example. For higher dimensional data
		this effect would only be more pronounced.

		\begin{figure}[t]
			\begin{floatrow}
				\capbfigbox{
					\input{figures/parzen_lls.tex}
				}{
					\caption{Parzen window estimates for a Gaussian evaluated on 6 by 6
						pixel image patches from the CIFAR-10 dataset. Even for small patches and a
						very large number of samples, the Parzen window estimate is far from the
						true log-likelihood.}
					\label{fig:parzen_windows}
				}
				\capbtabbox{
					\setlength{\tabcolsep}{.1em}
					\begin{tabular}{ccc}
						\toprule
						\textbf{Model} & Parzen est. [nat] \\
						\midrule
						Stacked CAE & 121 \\
						DBN & 138 \\
						GMMN & 147 \\
						Deep GSN & 214 \\
						Diffusion & 220 \\
						GAN & 225 \\
						\textbf{True distribution} & \textbf{243} \\
						GMMN + AE & {\color{red}282} \\
						$k$-means & {\color{red}313} \\
						\bottomrule
					\end{tabular}
					\vspace{.5cm}
				}{
					\caption{Using Parzen window estimates to evaluate various models trained on
						MNIST, samples from the true distribution perform worse than
						samples from a simple model trained with $k$-means.}
					\label{tab:results-parzen}
				}
			\end{floatrow}
		\end{figure}

		While the Parzen window estimate may be far removed from a model's true log-likelihood,
		one could still hope that it produces a similar or otherwise useful ranking when applied
		to different models. Counter to this idea, Parzen window estimates
		of the likelihood have been observed to produce rankings different from other estimates \citep{Bachman:2015}. More
		worryingly, a GMMN+AE \citep{Li:2015} is assigned a higher score than images from the training set (which
		are samples from the true distribution) when evaluated on MNIST (Table \ref{tab:results-parzen}).
		Furthermore it is relatively easy to exploit the Parzen window loss function to achieve even better results.
		To illustrate this, we fitted 10,000 centroids to the training data using $k$-means.
		We then generated 10,000 independent samples by sampling centroids with replacement. Note
		that this corresponds to the model in Equation~\ref{eq:lookup_table}, where the standard deviation of
		the Gaussian noise is zero and instead of training examples we use the centroids. We find
		that samples from this $k$-means based model are assigned a higher score than any other
		model, while its actual log-likelihood would be~$-\infty$.

	\section{Conclusion}
		We have discussed the optimization and evaluation of generative image models. Different
		metrics can lead to different trade-offs, and different evaluations favor different models.
		It is therefore important that training and evaluation match the target application.
		Furthermore, we should be cautious not to take good performance in one application as evidence of good
		performance in another application.

		An evaluation based on samples is biased towards models which overfit and therefore
		a poor indicator of a good density model in a log-likelihood sense, which favors models
		with large entropy. Conversely, a high likelihood does not guarantee visually pleasing
		samples. Samples can take on arbitrary form only a few bits from the optimum. It is therefore
		unsurprising that other approaches than density estimation are much more effective for image synthesis
		\citep{Portilla:2000,Dosovitskiy:2015,Gatys:2015}. Samples are in general also an unreliable
		proxy for a model's performance in applications such as classification or inpainting, as discussed in
		Section~\ref{sec:applications}.

		A subjective evaluation based on visual fidelity of samples is still clearly appropriate when the
		goal is image synthesis. Such an analysis at least has the property that the data
		distribution will perform very well in this task. This cannot be said about Parzen window
		estimates, where the data distribution performs worse than much less desirable
		models\footnote{In decision theory, such a metric is called an \textit{improper scoring
		function.}}. We therefore argue Parzen window estimates should be avoided for evaluating generative models,
		unless the application specifically requires such a loss function. In this case, we have
		shown that a k-means based model can perform better than the true density.
		To summarize, our results demonstrate that for generative models there is no one-fits-all
		loss function but a proper assessment of model performance is only possible in the
		the context of an application.

	\subsubsection*{Acknowledgments}

	The authors would like to thank Jascha Sohl-Dickstein, Ivo Danihelka, Andriy Mnih, and Leon Gatys for their valuable
	input on this manuscript.

	\bibliographystyle{iclr2016_conference}
	\bibliography{references}


\end{document}

%% file: figures/metrics.tex
\begin{tikzpicture}[scale=0.89,every node/.style={scale=0.89}]
	\def\x{3.9cm};

	\node at (0 * \x, 2.2cm) {\textsf{Data}};
	\node at (1 * \x, 2.2cm) {\textsf{KLD}};
	\node at (2 * \x, 2.2cm) {\textsf{MMD}};
	\node at (3 * \x, 2.2cm) {\textsf{JSD}};
	\node at (0 * \x, 0cm) {\includegraphics[width=3.8cm,natwidth=700,natheight=700]{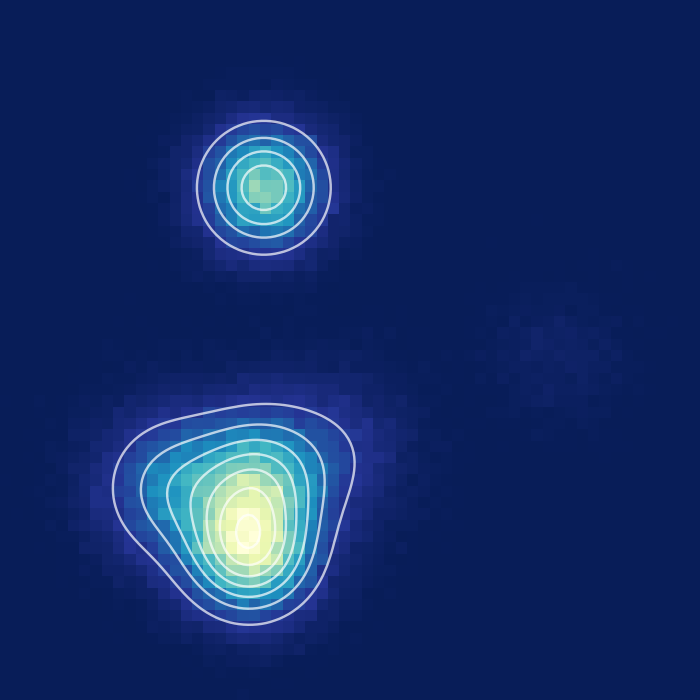}};
	\node at (1 * \x, 0cm) {\includegraphics[width=3.8cm,natwidth=700,natheight=700]{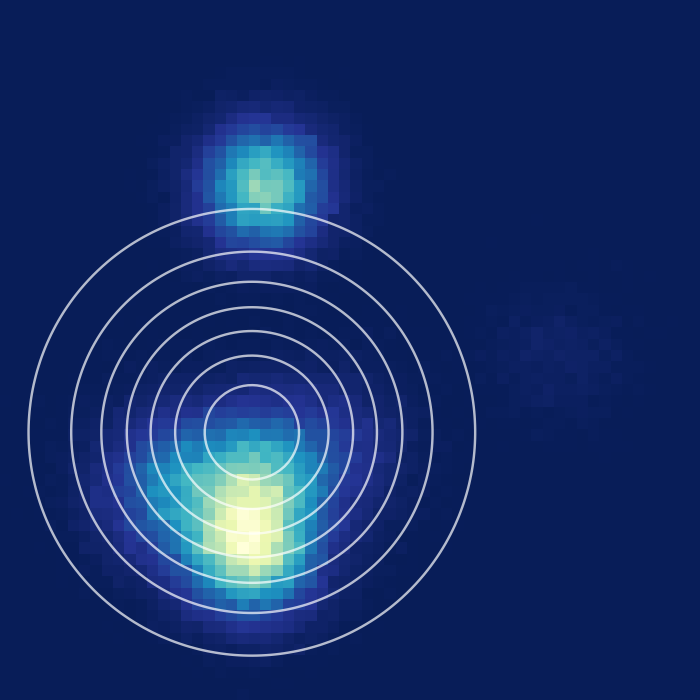}};
	\node at (2 * \x, 0cm) {\includegraphics[width=3.8cm,natwidth=700,natheight=700]{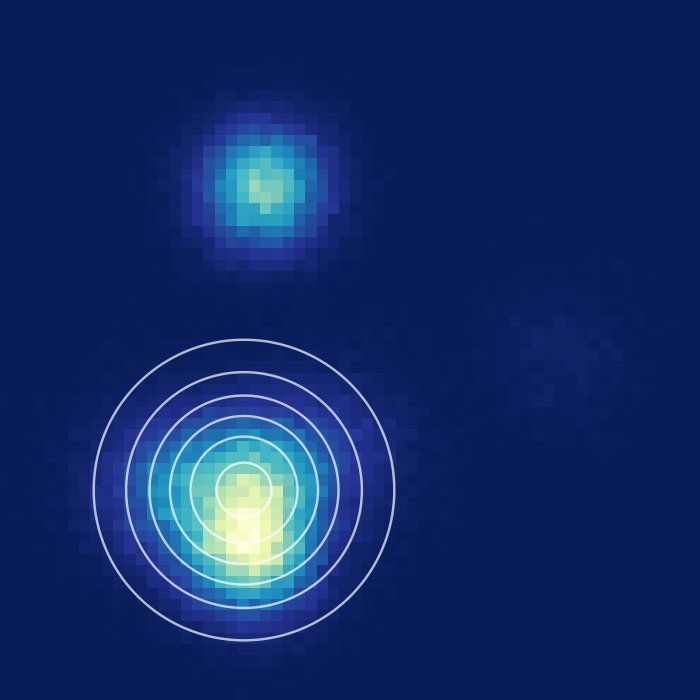}};
	\node at (3 * \x, 0cm) {\includegraphics[width=3.8cm,natwidth=700,natheight=700]{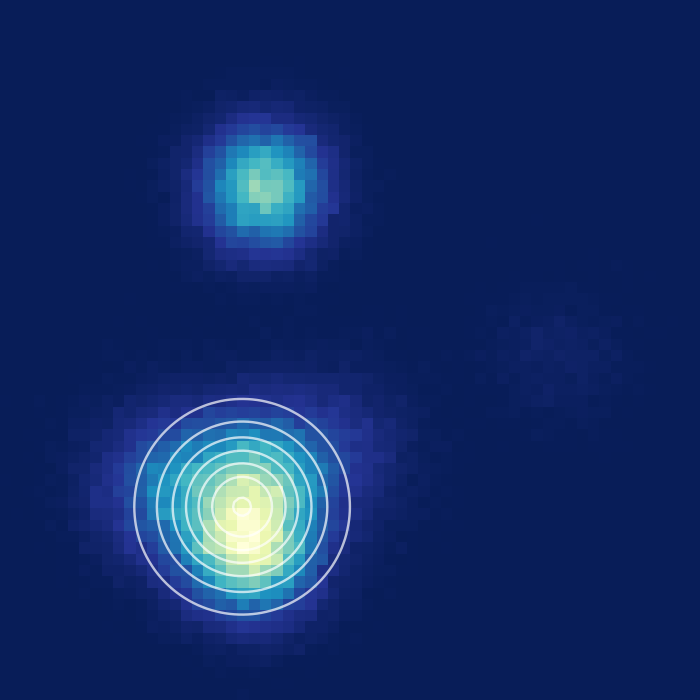}};
\end{tikzpicture}

%% file: figures/parzen_lls.tex
\begin{tikzpicture}
	\begin{axis}[
		scale only axis,
		width=4cm,
		height=3.5cm,
		at={(0.0cm, 0.0cm)},
		xmin=-1,
		xmax=7,
		ymin=0,
		ymax=240,
		xlabel={Number of samples},
		ylabel={Log-likelihood [nat]},
		axis on top,
		legend entries={\fontsize{8pt}{8pt}\selectfont Log-likelihood,\fontsize{8pt}{8pt}\selectfont Estimate},
		legend cell align=left,
		legend style={
			legend columns=-1,
			at={(1.1,1.2)},
			anchor=north east,
			font=\fontsize{8pt}{8pt}\selectfont\sffamily,
			draw=none,
			fill=none,
		},
		xlabel near ticks,
		ylabel near ticks,
		xtick={0,1,2,3,4,5,6},
		xtick scale label code/.code={},
		ytick={0,40,80,120,160,200,240},
		xticklabel={\pgfmathprintnumber[precision=4]{\tick}},
		yticklabel={\pgfmathprintnumber[precision=4]{\tick}},
		zticklabel={\pgfmathprintnumber[precision=4]{\tick}},
		xticklabels={$10^1$,$10^2$,$10^3$,$10^4$,$10^5$,$10^6$,$10^7$},
		label style={font=\fontsize{8pt}{8pt}\sansmath\sffamily\selectfont},
		tick label style={font=\fontsize{8pt}{8pt}\sansmath\sffamily\selectfont},
		axis x line=bottom,
		axis y line=left]
		\addplot+[no marks, line width=2pt, black, mark options={solid}] coordinates {
			(-1, 223.566)
			(7, 223.566)
		};
		\addplot+[solid, line width=2pt, red, mark=*, mark options={solid}] coordinates {
			(0, 14.2658158387)
			(1, 47.1105322147)
			(2, 65.434466924)
			(3, 75.2333263366)
			(4, 80.5503094447)
			(5, 83.7255376223)
			(6, 85.6198111307)
		};
	\end{axis}
\end{tikzpicture}